\documentclass[preprint,12pt]{elsarticle}

\usepackage{amsmath,amssymb,amsfonts,amsthm,mathrsfs,xspace}

\usepackage{array,tabularx,booktabs,multirow,makecell,threeparttable}
\newcolumntype{Y}{>{\raggedright\arraybackslash}X}
\newcolumntype{L}[1]{>{\raggedright\arraybackslash}p{#1}}

\usepackage{graphicx}
\usepackage[dvipsnames,table]{xcolor}

\usepackage{algorithm}
\usepackage{algorithmicx}
\usepackage{algpseudocode}

\usepackage{listings}
\usepackage{siunitx}
\usepackage[title]{appendix}

\usepackage{pgfplots}
\usepgfplotslibrary{groupplots}
\pgfplotsset{compat=1.17} % use 1.18 if supervisor has recent TeX

\usepackage{enumitem}

\newcommand{\mymethod}{ActLumos}

\makeatletter
\DeclareRobustCommand\onedot{\futurelet\@let@token\@onedot}
\def\@onedot{\ifx\@let@token.\else.\null\fi\xspace}
\def\eg{\emph{e.g}\onedot} 
\def\ie{\emph{i.e}\onedot} 
 
 \def\vs{\emph{vs}\onedot}
 
\def\etal{\emph{et al}\onedot}
\makeatother

\begin{document}

\begin{frontmatter}



\title{Seeing in the Dark: A Teacher–Student Framework for Dark Video Action Recognition via Knowledge Distillation and Contrastive Learning}

%\author{Sharana Dharshikgan Suresh Dass} %% Author name

%\affiliation{organization={School of Information Technology, Monash University},%Department and Organization
%country={Malaysia}}

%\author{Hrishav Bakul Barua}
%\affiliation{organization={Faculty of Information Technology, Monash University},
%country={Australia \& Malaysia}}
%\affiliation{organization={Robotics \& Autonomous Systems Lab, TCS Research},
%country={India}}
            
%\author{Ganesh Krishnasamy}
%\affiliation{organization={School of Information Technology, Monash University},%Department and Organization
%country={Malaysia}}

%\author{Raveendran Paramesran} %% Author name

%\affiliation{organization={Dept of Electrical Engineering, University of Malaya},%Department and Organization
%country={Malaysia}}

%\author{Raphael C.-W. Phan} %% Author name

%\affiliation{organization={School of Information Technology, Monash University},%Department and Organization
%country={Malaysia}}

\author[monash]{Sharana Dharshikgan Suresh Dass\fnref{scholar}}
\author[monash-aus,tcs]{Hrishav Bakul Barua\fnref{funding}}
\author[monash]{Ganesh Krishnasamy\fnref{corres}}

\author[malaya]{Raveendran Paramesran}
\author[monash]{Raphael C.-W. Phan}

\address[monash]{School of Information Technology, Monash University, Malaysia}
\address[monash-aus]{Faculty of Information Technology, Monash University, Australia}
\address[tcs]{Robotics \& Autonomous Systems Lab, TCS Research, India}

\address[malaya]{Dept of Electrical Engineering, University of Malaya, Malaysia}

\fntext[scholar]{This research is supported by the Global Research Excellence Scholarship, Monash University, Malaysia.}
\fntext[funding]{This research is also supported, in part, by the Global Excellence and Mobility
Scholarship (GEMS), Monash University, Malaysia \& Australia.}
\fntext[corres]{Corresponding author}

%% Abstract
\begin{abstract}
Action recognition in dark or low-light (under-exposed) videos is a challenging task due to visibility degradation, which can hinder critical spatiotemporal details. This paper proposes \mymethod, a teacher–student framework that attains single-stream inference while retaining multi-stream level accuracy. The teacher consumes dual stream inputs, which include original dark frames and retinex-enhanced frames, processed by weight-shared \textit{R(2+1)D-34} backbones and fused by a Dynamic Feature Fusion (DFF) module, which dynamically re-weights the two streams at each time step, emphasizing the most informative temporal segments. The teacher is also included with a supervised contrastive loss (\textit{SupCon}) that sharpens class margins. The student shares the \textit{R(2+1)D-34} backbone but uses only dark frames and no fusion at test time. The student is first pretrained with self-supervision on dark clips of both datasets without their labels and then fine-tuned with knowledge distillation from the teacher, transferring the teacher’s multi-stream knowledge into a single-stream model. Under single-stream inference, the distilled student attains state-of-the-art accuracy of 96.92\% (Top-1) on ARID V1.0, 88.27\% on ARID V1.5, and 48.96\% on Dark48. Ablation studies further highlight the individual contributions of each component, \ie, DFF in the teacher outperforms single/static fusion, knowledge distillation (KD) transfers these gains to the single-stream student, and two-view spatio-temporal SSL surpasses spatial-only or temporal-only variants without increasing inference cost. The official website of this work is available at:
https://github.com/HrishavBakulBarua/ActLumos 
\end{abstract}

\begin{graphicalabstract}
\centering
  \includegraphics[width=\linewidth]{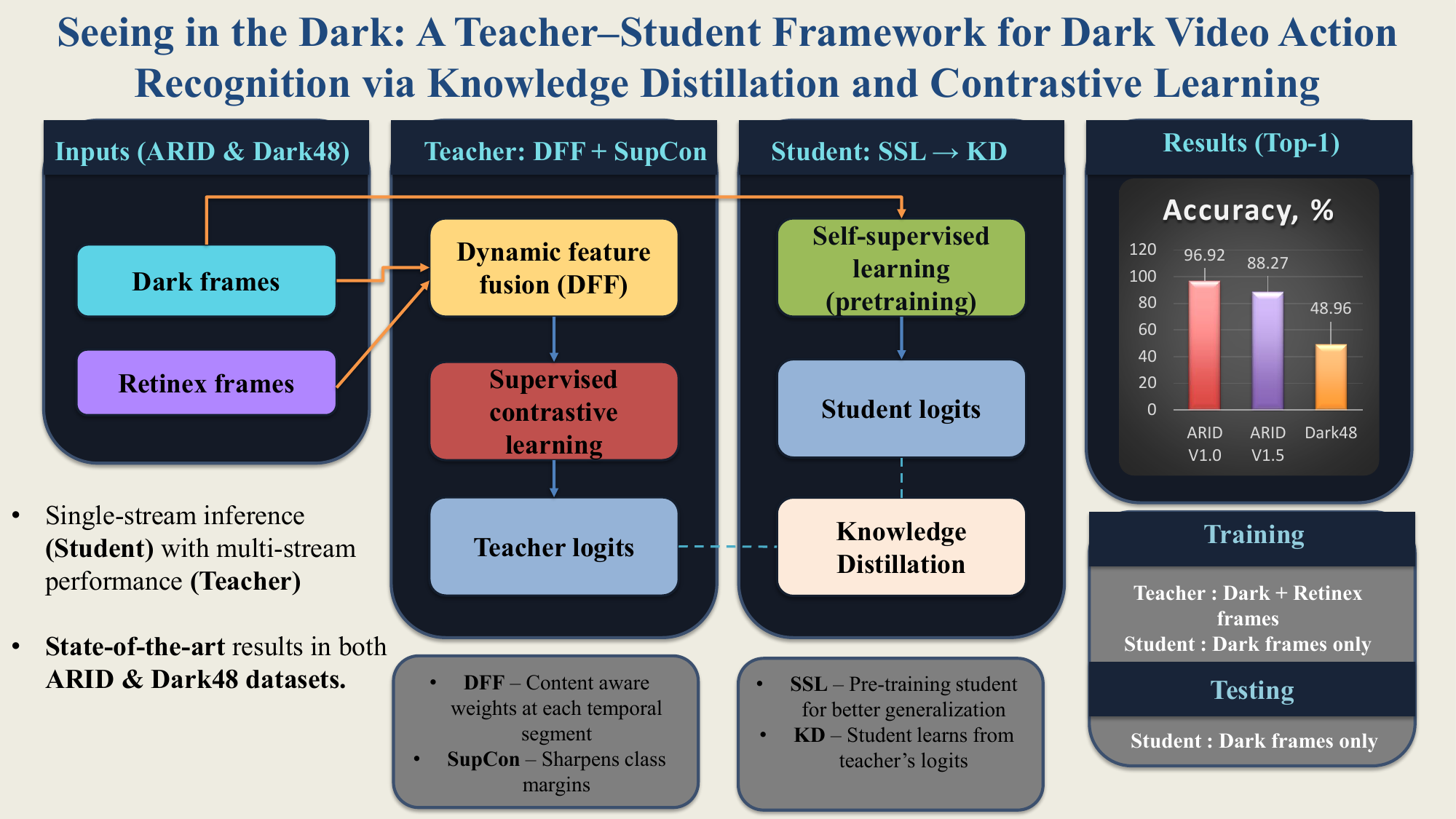}
\end{graphicalabstract}

\begin{highlights}
\item Teacher–student framework for low-light video action recognition.
\item Dynamic feature fusion reweights dark and retinex features at each temporal segment. 
\item Supervised contrastive learning sharpens class margins in the teacher.
\item Self-supervised pretraining plus distillation boosts student accuracy.
\item Single-stream inference delivers multi-stream-level accuracy.
\end{highlights}

%% Keywords

\end{frontmatter}



\section{Introduction}\label{introduction}

Action recognition in videos plays a non trivial role in applications such as surveillance, human-machine interaction, and automatic video tracking~\cite{dass2025actnetformer,dass2023schatten}. Although this area has been profusely explored in the vision community, recognizing actions in low-light conditions, such as night-time, as well as under-exposed surveillance or poorly lit environments, remains an under-explored area with significant challenges and a visible gap. These conditions often lead to reduced visibility, amplified noise, and loss of spatial and temporal details, significantly hindering model performance.

Traditional action recognition methods, which rely on handcrafted features or shallow architectures, struggle in such challenging scenarios. While deep learning models like R(2+1)D~\cite{tran2018closer} and I3D~\cite{carreira2017quo} have advanced the field through spatiotemporal feature learning, they typically assume well-lit environments, limiting their effectiveness in dark video scenarios. The challenges of action recognition in dark videos have prompted researchers to explore various strategies to address the limitations posed by low-light environments.

For example, Chen \etal~\cite{chen2021darklight} proposed a dual-stream approach that combines dark frames and gamma-corrected frames to improve feature representation in low-light settings. While effective, their approach relies on static concatenation of features, potentially under-utilizing complementary information. Similarly, Singh \etal~\cite{singh2022action} introduced an Image Enhancement Module (IEM) with Zero-DCE to enhance dark frames and coupled it with advanced temporal modeling techniques, demonstrating significant improvements. However, these methods often involve fixed preprocessing steps or lack the flexibility to adaptively fuse features.

%In contrast, we propose a two-stage teacher–student framework in which the teacher consumes two inputs, \ie, dark and retinex-enhanced frames. This approach was inspired from \cite{chen2021darklight} as they have shown promising results when using dual-stream (dark+gamma enhanced) compared to single strem appraoch architecture for dark video action recogntion. Hence, dark and retinex-enhanced frames that we are used will provide complementary video represenatiation. As illustrated in Figure~\ref{fig:dark-retinex-examples}, the two streams are complementary, \ie, the dark frames preserve the original appearance structure, while the retinex-enhanced frames recover edges and textures that are suppressed in deep shadows.

\begin{figure}[t]
  \centering
  \includegraphics[width=\linewidth]{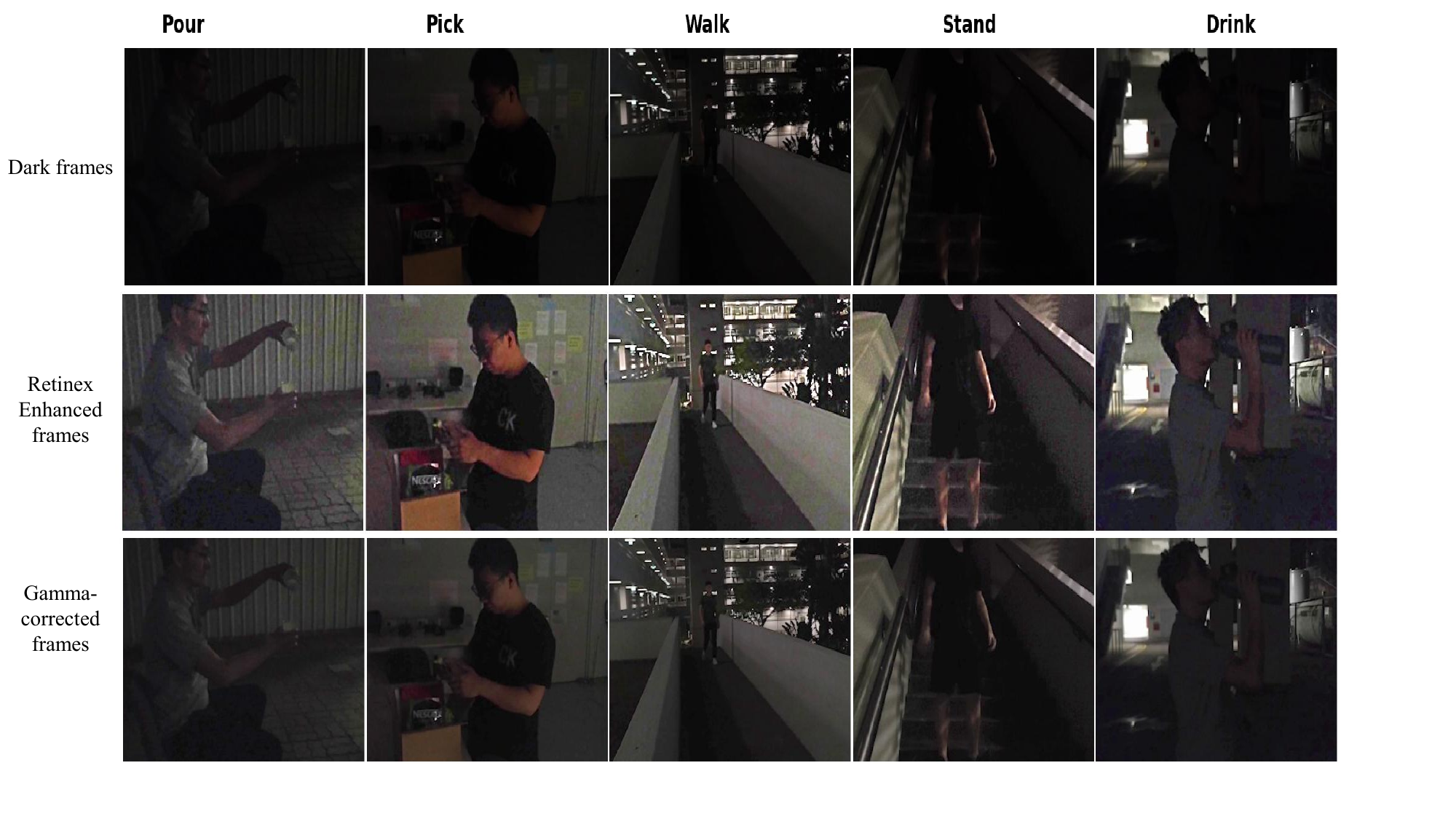}
  \caption{Examples of dark frames (top) and their retinex-enhanced counterparts (middle) and gamma-corrected frames, across actions (pour, pick, walk, stand, drink). Here the dark frames are from ARID dataset.}
  \label{fig:dark-retinex-examples}
\end{figure}

In contrast, we propose a two-stage teacher–student framework in which the teacher consumes two inputs, original dark frames and their retinex-enhanced counterparts. This design is inspired by~\cite{chen2021darklight}, which reports that dual-stream inputs (dark + gamma-enhanced) outperform single-stream architectures for action recognition in low-light. Instead of simple gamma correction, we adopt a retinex-based enhancement~\cite{guo2016lime}, which estimates a spatially varying illumination map and adaptively brightens dark regions while preserving edges and avoiding artifacts, yielding better quality. Thus, using both dark and retinex-enhanced frames provides complementary video representations. Figure~\ref{fig:dark-retinex-examples} shows the complementary dark and retinex streams and, alongside a gamma-corrected baseline, demonstrates that retinex reveals more detail with fewer artifacts than the gamma approach used in~\cite{chen2021darklight}.

%The two complementary streams are illustrated in Figure \ref{fig:dark-retinex-examples}.

\begin{figure}[t]
  \centering
  \includegraphics[width=\linewidth]{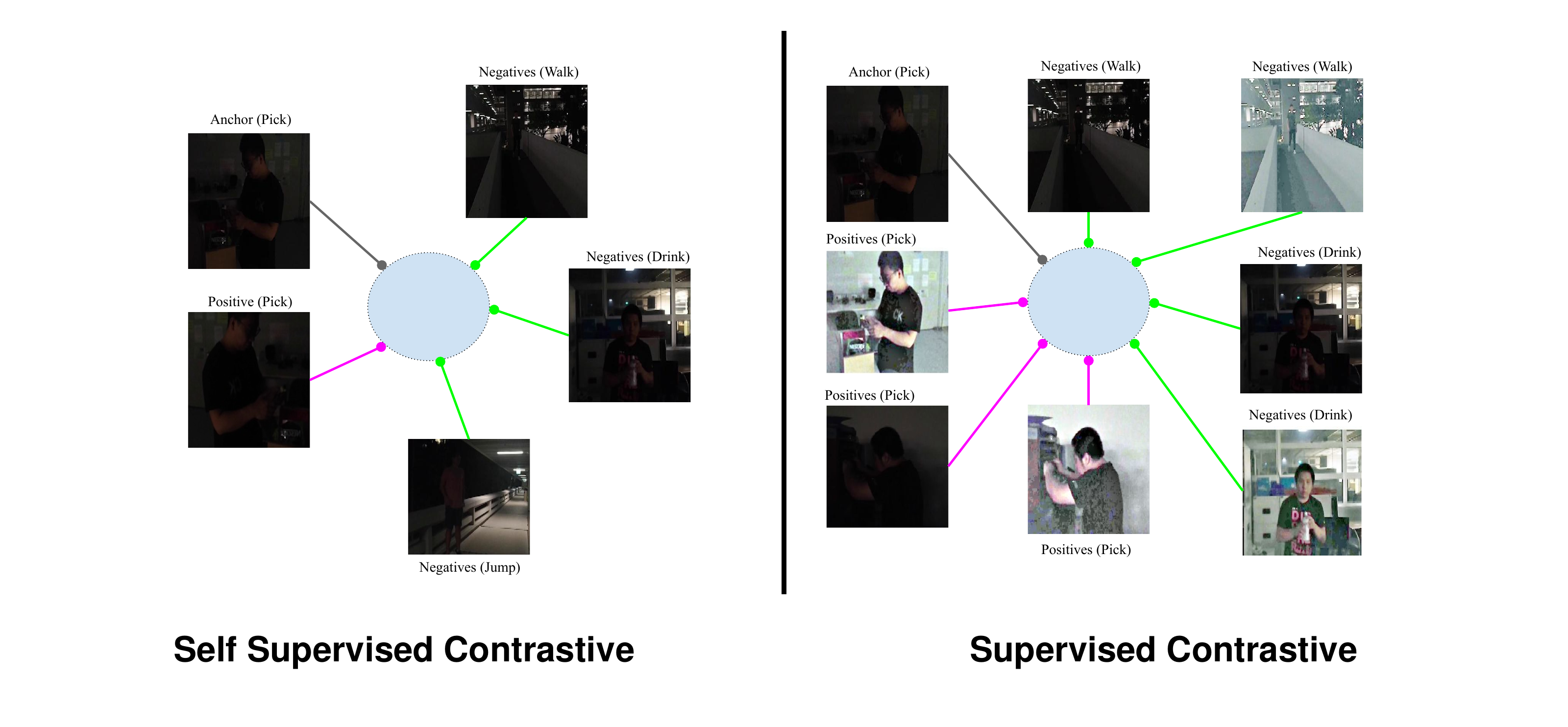}
  \caption{Self-supervised vs\ supervised contrastive learning. \textbf{Left} (self-supervised): the anchor clip (class \emph{Pick}) has only its own augmented view as a positive (pink edge), all other clips in the batch are treated as negatives (green). \textbf{Right} (supervised): with labels, every clip from the same class \emph{Pick} including dark and retinex views of different instances is a positive (pink), while clips from other classes are negatives (green).}
  \label{fig:supcon_self}
\end{figure}

 To effectively combine these two streams, we draw inspiration from the Attentional Feature Fusion (AFF) module~\cite{dai2021attentional}. However, AFF is image-centric (uses 2-D convolution kernels) and lacks temporal modeling. If applied directly to video inputs, AFF runs frame-by-frame without temporal context, thus losing the temporal dynamics. We therefore introduce Dynamic Feature Fusion (DFF), which adaptively re-weights dark and retinex-enhanced features at each temporal segment using 3-D convolution kernels that keep the temporal axis intact. This allows the network to favor the enhanced stream in darker moments and switch back to dark frames when lighting improves. The fusion gate learns weights at each timestep, so there is no preset preference for either stream. It is worth noting that the teacher processes both dark and retinex-enhanced streams with shared weights. The fused features produced by the DFF module are used to model long-range temporal dependencies with a BERT-style transformer model. 
 
 Furthermore, we deliberately use two contrastive objectives, but in different places in our proposed architecture as illustrated in Figure~\ref{fig:supcon_self}. In the teacher, we incorporate video-level supervised contrastive learning (SupCon). SupCon pulls all clips with the same action labels together and pushes different classes apart. It is used to discover the mutual information across all same-label clips while ignoring lighting-specific noise, so dark and enhanced clips of the same class end up close to each other in the feature space. 

In the second stage, a compact student that processes only dark video is first pre-trained by self-supervised contrastive learning on a large pool of unlabeled dark clips, thereby learning general low-light features without using labels. We then fine-tune the student by matching the teacher’s prediction via knowledge distillation while also training with cross-entropy on the ground-truth labels. It is worth noting that we do not run SupCon on the student during pretraining. We keep it label-free so it can learn broad view/temporal invariances from a larger pool of dark clips (\ie, ARID and Dark48) for better generalization. Labels for data are introduced later via KD and CE, where CE denotes the student's standard cross-entropy classification loss, and KD is the distillation loss from the teacher. Moreover, since the teacher already uses SupCon, applying it to the student would be redundant and offer little additional gain. Likewise, we do not use SSL on the teacher because, with labels and two streams, SupCon is the stronger objective for class separation and cross-view alignment. At test time, we deploy only the student; the teacher exists purely for training/distillation.

%Because retinex processing (second stream) and DFF disappear at inference, the student retains the teacher’s low-light robustness while running at lower computational cost.

The main contributions of this work are fourfold and summarized as follows:
\begin{itemize}
  \item \textbf{Teacher-student design.} A two-stage framework where a dual-stream teacher (dark + retinex) distills a compact single-stream student, enabling single-stream inference while retaining the teacher’s multi-stream-level accuracy.
  \item \textbf{Dynamic Feature Fusion (DFF).} A 3D-based per-timestep gating module that adaptively re-weights dark and retinex features based on content.
  \item \textbf{Supervised contrastive learning.} A SupCon objective in the teacher that treats dark/retinex views and same-class clips as positives and different-class clips as negatives, yielding lighting-robust, well-separated embeddings.
\item \textbf{Student training recipe.} Self-supervised pretraining using unlabeled dark videos from both ARID and Dark48 datasets, followed by knowledge distillation (KD) using the teacher’s prediction, and cross-entropy fine-tuning.

\end{itemize}

\section{Literature Review}

\subsection{Action Recognition in Dark Videos}

Action recognition is an important area in computer vision with applications in surveillance, autonomous driving, human-computer interaction, and sports analytics. Early methods focused on hand-engineered features~\cite{laptev2008learning}, while modern approaches employ 3D CNNs~\cite{tran2015learning}, two-stream architectures~\cite{simonyan2014two}, and Transformer-based models~\cite{dosovitskiy2020image}, achieving state-of-the-art results. However, low-light video recognition remains underexplored due to challenges like poor visibility and loss of discriminative details. Xu \etal~\cite{xu2021ARID} introduced ARID dataset for dark video action recognition. Hira \etal~\cite{hira2021delta} proposed a delta-sampling approach integrating ResNet and BERT while Singh \etal~\cite{singh2022action}  employed Zero-DCE with R(2+1)D, GCN, and BERT for spatio-temporal feature extraction. Chen \etal~\cite{chen2021darklight} developed DarkLight Networks using a dual-pathway structure with self-attention for feature fusion. Suman \etal~\cite{suman2023two} introduced a two-stream technique combining SCI-based image enhancement and GCN for temporal refinement. Tu \etal~\cite{tu2023dtcm} proposed the Dark Temporal Consistency Model (DTCM), an end-to-end framework optimizing both enhancement and classification. Wang \etal~\cite{wang2025advancing} proposed Modality Fusion Dark-to-Light (MFDL), a two-stage framework to simultaneously enhance the invisibility of poorly-lit videos with the help of diffusion model.

\subsection{Multi-Feature Fusion}

Multi-feature fusion has been extensively studied in the context of action recognition due to its ability to leverage complementary information from various modalities or feature streams~\cite{feichtenhofer2016convolutional, carreira2017quo}. By combining diverse inputs such as RGB, optical flow, depth, skeleton data, or audio, these methods address the inherent complexity of video data and improve model resilience in challenging conditions. Early works like Two-Stream CNNs~\cite{simonyan2014two} demonstrated the effectiveness of parallel networks for spatial and motion cues, while subsequent research integrated additional modalities (\eg, depth~\cite{yang2012recognizing} or skeleton data~\cite{li2010action}) to further boost performance. More recent architectures have explored sophisticated attention mechanisms and feature fusion strategies to handle scale variations, semantic inconsistencies, and long-range dependencies in videos~\cite{zhang2020multi,alamri2019audio}.

In the broader landscape of feature fusion techniques, Attentional Feature Fusion (AFF)~\cite{dai2021attentional} has gained prominence as a crucial method. AFF effectively addresses both cross-layer and same-layer fusion challenges, mitigating issues such as semantic inconsistency across feature maps and the need for comprehensive multi-scale context aggregation. Specifically, it introduces a local attention branch (via convolutional layers) and a global attention branch (via global pooling) to adaptively highlight critical features. In dark videos, DarkLight and DTCM embrace low-light enhancement plus recognition, but typically rely on fixed concatenation~\cite{chen2021darklight, tu2023dtcm}. MFDL~\cite{wang2025advancing} adds modality-fusion modules within a dual-pathway backbone to encourage interaction between dark and brightened streams.

In contrast to static concatenation inspired by AFF, our teacher employs a Dynamic Feature Fusion (DFF) module that adaptively re-weights dark and enhanced streams at each time step, enabling context-dependent fusion and yielding higher accuracy.

\subsection{Supervised contrastive learning (Supcon)}

Supervised Contrastive Learning (SupCon) was introduced by Khosla \etal~\cite{khosla2020supervised} as a supervised variant of batch contrastive learning that uses all same-class samples in the batch as positives for each anchor, while treating different-class samples as negatives. Compared to plain cross-entropy (CE), SupCon explicitly shapes the embedding space by pulling together class clusters and pushing apart others, typically improving margins, robustness to corruptions, and optimization stability.

A number of action-recognition frameworks have adopted supervised (or semi-supervised) contrastive objectives to stabilize spatiotemporal features and improve generalization. Shah \etal~\cite{shah2023multi} proposed a supervised contrastive framework that augments positives with synchronized views, improving viewpoint robustness for action classification. Hirata \etal~\cite{hirata2021making} showed that supervised contrastive training yields more discriminative and stable video features and improves robustness under common corruptions. Seifi \etal~\cite{seifi2024ood} leveraged SupCon-trained representations for better out-of-distribution detection, indicating improved class separation that also benefits downstream video classification. Li \etal~\cite{li2022targeted} proposed Targeted Supervised Contrastive (TSC) learning, which modifies SupCon to emphasize minority classes and calibrate inter-class margins, thereby preventing minority-class collapse and promoting more uniform class centers under imbalance.

In our framework, SupCon is applied only in the teacher, not deployed in student. We compute the SupCon loss on the per-stream clip embeddings treating all same-class samples in the batch as positives and the cross-stream counterpart of the same video as an additional positive, whereas the negative pair is when the videos are different, regardless of whether they come from the dark or the enhanced stream. This setup pulls together dark and retinex views of the same action, promoting lighting invariance, while simultaneously pushing apart different classes to sharpen decision margins.

\subsection{Knowledge Distillation}

Knowledge Distillation (KD) was popularized by Hinton \etal~\cite{hinton2015distilling} as “distilling the knowledge in a neural network”. A high-capacity teacher produces softened targets via a temperature-scaled softmax, and a student is trained to match them using KL divergence or cross-entropy (CE). Subsequent work expanded on what is transferred. Born-Again~\cite{furlanello2018born} Networks showed that even capacity-matched students can surpass their teachers through iterative KD. VID~\cite{ahn2019variational} transfers information by maximizing mutual information between teacher and student feature maps. RKD~\cite{park2019relational} transfers relations (pairwise distances/angles) among samples.  

A number of action-recognition papers have adopted knowledge distillation into their framework, such as DistInit~\cite{girdhar2019distinit} that trains video encoders from image teachers, learning spatiotemporal features without video labels (image→video KD). The work introduced by Kim \etal~\cite{kim2021efficient} improves efficient action recognition by propagating knowledge to lighter networks under frame selection, while the work proposed by Dai \etal~\cite{dai2021learning} transfers temporal relations from auxiliary modalities during training to strengthen RGB detectors. D3D~\cite{stroud2020d3d} distills a flow-based teacher into an RGB-only 3D CNN so motion cues are learned without computing flow at test time. Prior KD frameworks for action recognition typically compress heavy 3D models, distill motion/flow into RGB, or collapse multimodal teachers into single-stream students. 

In contrast we keep the teacher and student capacity matched (both use R(2+1)D-34). 
The teacher’s only advantage is its input; it sees both dark and retinex streams, fused by DFF, and is trained with CE and SupCon. We then distill this teacher into a single-stream student, trained with CE and KD, that at inference uses only the dark frames.

\subsection{Self-supervised learning}

Self-supervised learning (SSL) learns visual encoders from unlabeled data by creating “two views” of the same sample and driving their representations together. Contrastive methods such as Momentum Contrast (MoCo)~\cite{he2020momentum} and SimCLR~\cite{chen2020simple} established the modern recipe for image SSL, showing that strong augmentations and temperature-scaled objectives yield transferable features. Extending SSL to videos requires temporal view construction in addition to spatial crops/photometric jitter. CVRL~\cite{Qian_2021_CVPR} and VideoMoCo~\cite{pan2021videomoco} showed that forming two clips from the same video with different spatiotemporal samplings (frame rate/clip start/drop frames) consistently outperforms spatial-only setups for downstream action recognition. 

Inspired by these works, we also include SSL in our approach; however, we only include it in our student model. Each view receives its own random crop/resize/flip as a form of spatial augmentation, and the two views use different temporal samplings (\eg, slower \vs faster frame rate), following best practice from~\cite{Qian_2021_CVPR, pan2021videomoco} to learn robustness to timing and apparent speed. We first pre-train the student with SSL on dark clips (no labels), then fine-tune it with KD from the dual-stream teacher.

\begin{figure*}[t]
    \centering
    \includegraphics[width=\textwidth]{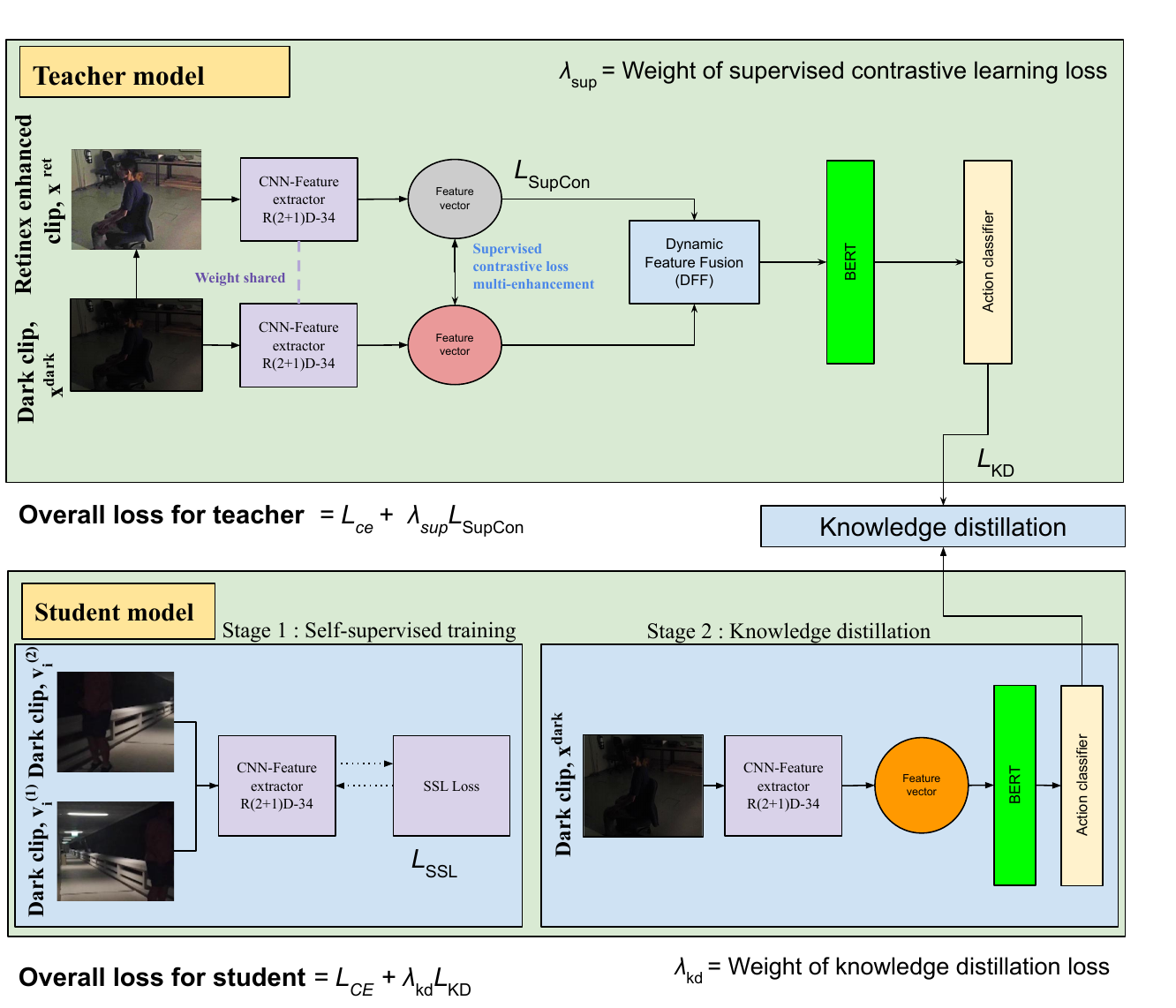}
    \caption{The framework for the proposed \textbf{\mymethod} approach.}
    \label{fig:arc-figure}
\end{figure*}

\section{Methodology}

\subsection{Overview}

This section details the complete learning pipeline, which incorporates a dual-stream \textbf{teacher} with a compact \textbf{student}.  
The teacher exploits both the original dark frames and retinex-enhanced frames, fusing them per temporal segment with our Dynamic Feature Fusion (DFF) block. Moreover, the teacher is also trained with a supervised contrastive loss. In parallel, the student uses the same backbone family (R(2+1)D-34) but is trained as a single-stream model. First, the student is pretrained in a self-supervised contrastive learning setup using the original dark video clip. We remove the labels from the ARID and Dark48 datasets and use all the clips as unlabeled data so that the student learns robust representations and generalizes better. Following the SSL pretraining, the student is then fine-tuned on the labeled target dataset (ARID or Dark48) by knowledge distillation from the frozen teacher, using temperature-scaled KL divergence between the teacher’s and student’s outputs combined with standard cross-entropy. This allows the student to inherit the teacher’s fusion knowledge and to sharpen class boundaries. Importantly, the student sees only dark frames and does not require (retinex enhanced frames or DFF) at inference, making it efficient for real-world low-light deployment. This yields multi-stream-level accuracy with single-stream complexity.

\subsection{Notation and Preliminaries}
\label{ssec:np}
A dark video clip is denoted $x^{\text{dark}} \in \mathbb{R}^{3 \times L \times H \times W}$,
where the three channels are RGB and $L,H,W$ are the number of frames, height, and width, respectively.
The retinex-enhanced stream is given by
$x^{\text{ret}} = R\!\left(x^{\text{dark}}\right)$, with $R(\cdot)$ a single-scale Retinex operator.
Both streams pass through a \emph{shared} 3-D backbone (R(2+1)D-34) followed by spatial
global-average pooling, resulting in the following feature representations:

\[
f^{\text{dark}} = h\!\bigl(x^{\text{dark}}\bigr)\in\mathbb R^{C\times T\times{H}\times{W}},
\qquad
f^{\text{ret}} = h\!\bigl(x^{\text{ret}}\bigr)\in\mathbb R^{C\times T\times{H}\times{W}}
\]
where \(h(\cdot)\) denotes the backbone, \(C = 512\) channels, and \(T\) = time steps after temporal downsampling (typically \(T = 8\)). The two feature tensors serve as inputs to the Dynamic Feature Fusion (DFF) block, as described in Section \ref{ssec:dff}. 

% These two tensors are flattened (spatial Global Average Pooling GAP) and $\ell_2$‑normalised to give one vector per view,
% The two tensors are processed using spatial Global Average Pooling (GAP) and subsequently $\ell_2$-normalized, resulting in a single feature vector representation for each view,
% \[
% z_i^{\text{dark}},\;z_i^{\text{ret}}\in\mathbb R^{C},
% \]
%  which will be used to form the 16‑vector set $\mathcal Z$ used by the supervised
% contrastive loss in Section ~\ref{ssec:supcon}.

The two tensors are processed using spatial Global Average Pooling (GAP) and subsequently $\ell_2$-normalized, yielding a single feature vector for each view, $z_i^{\text{dark}}, z_i^{\text{ret}} \in \mathbb{R}^{C}$. These vectors collectively form the 16-vector set $\mathcal{Z}$, which is employed in the supervised contrastive loss described in Section~\ref{ssec:supcon}.

\subsection{Supervised Contrastive Learning}
\label{ssec:supcon}
.
We extend the supervised contrastive learning (SupCon) \cite{khosla2020supervised} to our dual-stream teacher, enabling it to learn features that are both \emph{lighting-invariant and class-discriminative}. Each mini-batch is \emph{class balanced}: we randomly pick \(n_c\) distinct action labels and then sample \(n_v\) different clips per label. In all experiments, we use \(n_c=4\) classes and \(n_v=2\) clips per class, giving \(B=16\) vectors while loading only eight physical video tensors. Every clip appears twice, its dark view and its retinex view, so the batch contains:
\[
B = 2\,n_c n_v
\]

\paragraph{Positive and negative sets.}
Let
$\mathcal Z=\{z_1,\dots,z_{B}\}\subset\mathbb R^{C}$
be the $\ell_2$‑normalised features after the backbone
($C=512$) and let $y_i\in\{1,\dots,K\}$ be the action
label of $z_i$.
For an \emph{anchor} index $i$ we define:
\begin{equation}
\begin{aligned}
\mathcal P(i) &= \{\ \underbrace{p}_{\text{other index}}\neq i
                 \mid \underbrace{y_p}_{\text{its label}}=y_i \ \},\\
\mathcal N(i) &= \{\ \underbrace{a}_{\text{other index}}\neq i
                 \mid y_a \neq y_i \ \}.
\end{aligned}                 
\end{equation}

\begin{description}
\item[$\mathcal P(i)$:] \emph{all other} vectors in the batch that share the \textbf{same class}
      label as the anchor (positives),
\item[$\mathcal N(i)$:] every vector whose label is \textbf{different}
      from the anchor’s (negatives).
\end{description}

\begin{table}[t]
\centering
\renewcommand{\arraystretch}{1.05}
\begin{tabular}{|c|l|c|c|}
\hline
\textbf{index} & \textbf{view}            & \textbf{class $y$} & \textbf{role} \\
\hline
\rowcolor{lightgray}0  & dark (clip 1)        & A & anchor   \\
1  & retinex (clip 1)     & A & positive \\
2  & dark (clip 2)        & A & positive \\
3  & retinex (clip 2)     & A & positive \\
\hline
4  & (dark/retinex, other class) & B, C, D & \multirow{4}{*}{negative} \\

$\vdots$ & $\vdots$                 & $\vdots$ & \\

16 & (dark/retinex, other class) & B, C, D & \\
\hline
\end{tabular}
\caption{Example batch with $n_c{=}4$ classes and $n_v{=}2$ clips per class ($B{=}16$ vectors). 
With anchor $i{=}0$ (dark view of class A, clip 1), indices $\{1,2,3\}$ are positives, while indices $\{4,5,\ldots,16\}$ (from other classes) are negatives.}
\label{tab:batch_example}
\end{table}

%With anchor $i=0$ (dark view of clip 1), we have 
%$\mathcal{P}(0)=\{1,2,3\}$ (three positives), while the remaining twelve indices from the other three classes serve as negatives. In other words, every anchor treats \emph{all} dark and retinex views of its own class as positives, and all views from other classes as negatives. With $n_c=4$ classes and $n_v=2$ clips per class, this yields $|\mathcal{P}(i)|=3$ positives and $|\mathcal{N}(i)|=12$ negatives, providing a strong, balanced contrastive signal even at small batch size. 
Thus \(|\mathcal P(i)| = 2n_v-1\) and \(|\mathcal N(i)| = B-1-|\mathcal P(i)|\), with
\(n_c{ = } 4,n_v{ = } 2\) this gives \(3\) positives and \(12\) negatives (e.g., for anchor
\(i{=}0\): \(\mathcal P(0) = \{1,2,3\}\)). The supervised contrastive loss for that anchor is as follows: 

\begin{equation}
\mathcal{L}^{\text{SupCon}}_i
= -\frac{1}{\lvert \mathcal{P}(i)\rvert}
  \sum_{p \in \mathcal{P}(i)}
  \log \frac{\exp(z_i^{\top} z_p/\tau)}
               {\sum_{a \neq i} \exp(z_i^{\top} z_a/\tau)}.
\label{eq:supcon_anchor}
\end{equation}

where \(\tau\) is the temperature (\(\tau = 0.1\)), $z_i$ = anchor representation, 
$z_p$ = positive samples (same class as $z_i$), 
$z_a$ = all other samples in the batch.
Following the original formulation in \cite{khosla2020supervised},
we \emph{sum} the anchor losses over the batch
\begin{equation}  
\mathcal L_{\text{SupCon}}
\;=\;
\frac{1}{B}\sum_{i=0}^{B}
\mathcal L^{\text{SupCon}}_i .
\label{eq:supcon_batch}
\end{equation}

Finally, the teacher is trained with a cross-entropy loss, together with the supervised contrastive loss:

\begin{equation}
\mathcal L_{\text{Teacher}}
\;=\;
\mathcal L_{\text{CE}}
\;+\;
\lambda_{\text{sup}}\,
\mathcal L_{\text{SupCon}}
,
\qquad
\lambda_{\text{sup}} = 0.1 .
\label{eq:teacher_total}
\end{equation}

\begin{figure}[t]
    \centering
    \includegraphics[width=\linewidth]{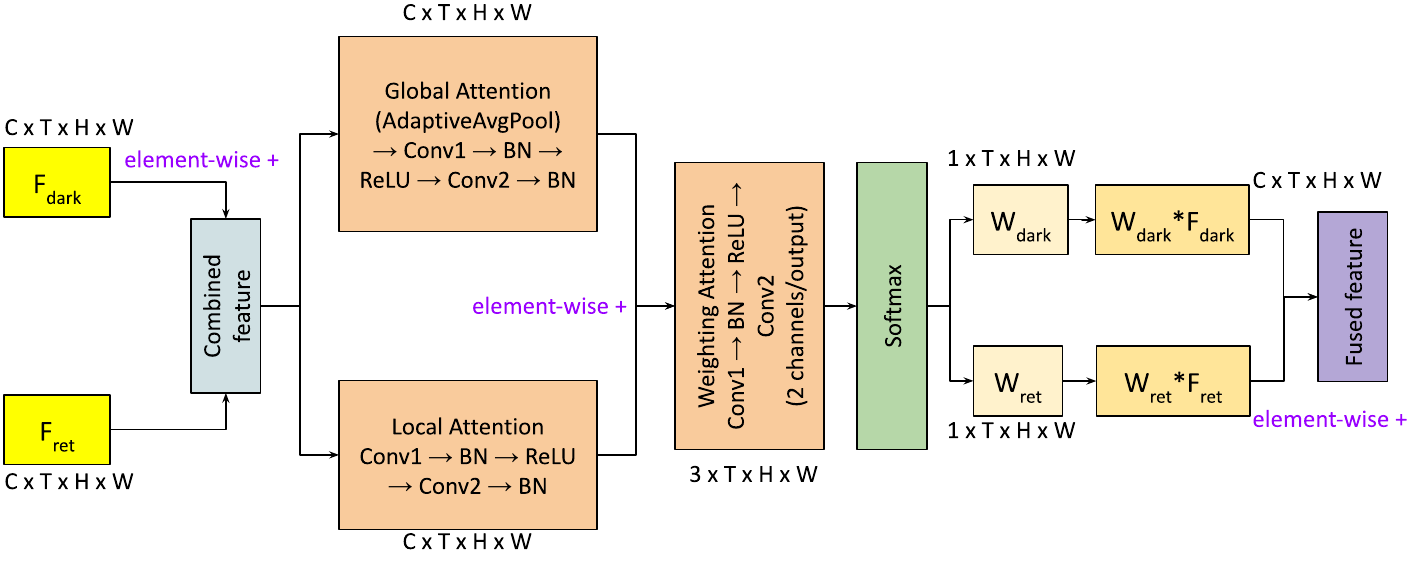}
    \caption{Illustration of the proposed Dynamic Feature Fusion (DFF) module. At each temporal step, it adaptively weighs the dark and retinex features to select the most informative representation.}
    \label{fig:dff}
\end{figure}

\subsection{Dynamic Feature Fusion (DFF)}
\label{ssec:dff}

The illumination within a scene is often dynamic and non-uniform; for instance, a subject may transiently traverse a well-lit area before receding into profound shadow, or the entire scene may oscillate between low and moderate light levels. Under conditions of mild darkness, the dark stream typically preserves structural and motion information with greater reliability than retinex-based enhancement. However, in regions of extreme low light, the dark signal degrades, and the retinex output becomes more informative by revealing low contrast edges. A single, static fusion weight is inherently inadequate to adapt to these rapid, localized transitions in informational salience. Consequently, a significant portion of the complementary data offered by each stream remains underutilized.

Our proposed Dynamic Feature Fusion (DFF) module, as illustrated in Figure~\ref{fig:dff}, addresses the time-varying informativeness of the two streams by predicting an adaptive weight for every temporal segment. At each latent time step, it examines the paired feature vectors from the two streams, judges which one is richer for that moment, and outputs a soft gate that balances their contributions. When the dark features are attenuated by heavy darkness, the gate leans toward the retinex features, whereas when the dark stream is already informative, the gate favors it instead. This content-aware, segment-wise weighting ensures the network consistently focuses on the strongest available information in either stream throughout the clip.

% \textcolor{red}{Hard to understand...Lighting conditions fluctuate even within the same clip: some segments
% of the \emph{dark} stream are already informative, whereas darker
% segments benefit strongly from the \emph{retinex} view.} For instance, a passer-by \textcolor{red}{(person?)} may briefly enter a pool of light, then move back into deep shadow, or a scene may alternate between dim and moderately lit areas. In mildly dark portions of a clip, the dark stream still preserves reliable structure and motion information better than a retinex version, while in really dark portions, the dark signal collapses, and the retinex view becomes more informative by pulling out hidden edges and contrast. A single, fixed fusion weight cannot capture these rapid shifts, so complementary information is left unused.

\paragraph{Gate prediction}
For each latent time step \(t\), we take the two
\(C\)-dimensional feature vectors \(f_t^{\text{dark}}\) and \(f_t^{\text{ret}}\)
(defined in Section \ref{ssec:np}), concatenate them, and pass the result through
a lightweight 3‑D MLP:

\begin{equation}
\begin{aligned}
g_t = \phi\!\bigl([\,f_t^{\text{dark}},\,f_t^{\text{ret}}]\bigr)\in\mathbb{R}^{2},
\qquad
[w_t^{\text{dark}},\,w_t^{\text{ret}}]=\texttt{softmax}(g_t).
\end{aligned}
\end{equation}
Then, the fused feature for that segment is given by,
\begin{equation}
\begin{aligned}
f_t = w_t^{\text{dark}}\,f_t^{\text{dark}} + w_t^{\text{ret}}\,f_t^{\text{ret}},
\quad
\textstyle w_t^{\text{dark}}+w_t^{\text{ret}}=1.
\end{aligned}
\end{equation}

%\textcolor{red}{Stacking all $f_t$ yields $F\in\mathbb{R}^{C\times T}$}, which is \textcolor{red}{projected} and passed to the temporal transformer. The resulting \texttt{[CLS]} embedding passes through a fully connected head to produce class logits, which are trained with cross-entropy and the supervised contrastive loss described in Section \ref{ssec:supcon}.

The fused features, $f_t$, over time are stacked into a sequence
$F=[f_1,\ldots,f_T]\in\mathbb{R}^{T\times C}$, where $T$ is the number of temporal segments and $C$ is the channel dimension. A learnable temporal positional embedding is added to each fused feature and a learnable [CLS] token is prepended, producing $B\in\mathbb{R}^{(T+1)\times C}$ as input to the temporal
transformer; the [CLS] output is used for classification.

\subsection{Stage 1 — Self-Supervised Contrastive Pre-training of the Student}
To promote strong generalization, we first pretrain the R(2+1)D-34 student on unlabeled dark videos to learn stable spatiotemporal representations and encourage consistency between the two views of the same clip. Then, we fine-tune it via knowledge distillation (KD) on the target dataset. We use all raw clips from \textbf{ARID} and \textbf{Dark48} (labels discarded) in a two-view instance-contrastive setup. Mixing these datasets enlarges the pool of low-light content (different cameras, noise patterns, and motion statistics), improving robustness without creating label conflicts.

\begin{enumerate}[leftmargin=1.6em,itemsep=2pt]
  \item \textbf{View 1 ($v_i^{f}$):} Frames sampled at a higher rate. We use random horizontal flip, random scale jitter, and random crop as a form of spatial augmentation and apply them to all frames in this view.
  \item \textbf{View 2 ($v_i^{s}$):} Frames sampled at a lower rate. Similar spatial augmentation is similar to view 1, but the random parameters are re-sampled independently for this view and then applied consistently to all its frames.
\end{enumerate}
These two views share the same underlying action but differ in appearance and pacing, encouraging the encoder to focus on motion and pose cues that survive common variations in dark videos.

We form each self-supervised mini-batch by sampling $B_u$ unlabeled \emph{dark} clips from ARID and Dark48 (labels discarded). For every clip $x_i$ we create two augmented views: a \emph{fast} view $v_i^{f}$ sampled at a higher frame rate with spatial augmentation, and a \emph{slow} view $v_i^{s}$ sampled at a lower frame rate with the same family of spatial augmentations but independent parameters. Let $g(\cdot)$ be the student encoder. It produces feature embeddings
$z_i^{f}=g(v_i^{f})$ and $z_i^{s}=g(v_i^{s})$. The pair $(z_i^{f},\,z_i^{s})$ is the \emph{positive} for clip $i$. For the rest $B_u - 1$ videos, $z^{(f)}_i$ and $z^{(k)}_q$ form negative pairs, where the representation of the $q$-th video can come from either of the view (i.e., $k \in \{f,s\}$). Since negatives come from different videos with distinct content, the contrastive objective
increases the similarity between the two augmented \emph{dark} views of the same clip
(anchor and its counterpart) while decreasing similarity to all views from other clips. We achieve this using the contrastive loss $\mathcal{L}_{\text{SSL}}^{(i)}$, adapted from the InfoNCE formulations in~\cite{chen2020simple,singh2022action, dass2025actnetformer}, as detailed below.

\begin{equation}
\scalebox{0.9}{$
\mathcal{L}_{\text{ssl}}(z^{f}_i, z^{s}_i) = -\log
\frac{
    p\big(z^{f}_i,\, z^{s}_i\big)
}{
    p\big(z^{f}_i,\, z^{s}_i\big) + 
    \sum_{\substack{q=1\\q \ne i}}^{B} 
    \sum_{k \in \{f,s\}} p\big(z^{f}_i,\, z^{k}_q\big)
}
$}
\label{eq:ssl_loss}
\end{equation}

where, \( p(u, v) \) = $\exp \left( \frac{u^\top v}{\| u \|_2 \| v \|_2} /\tau \right) $ represents the exponential of the cosine similarity measure between vectors \( u \) and \( v \), where \( \tau \) denotes the temperature hyperparameter. The final contrastive loss is computed over all positive pairs $(z_i^{f}, z_i^{s})$, where $z_i^{f}$ and $z_i^{s}$ are the embeddings of the \emph{fast}- and \emph{slow}-rate augmented views of the same dark clip $x_i$. The objective increases the similarity between these two views of $x_i$ while simultaneously reducing similarity to all views from \emph{other} clips in the batch, regardless of whether they are fast- or slow-rate views.

%--------------------------------------------------------------
\paragraph{Hand-off to Stage 2}
After SSL, we append a linear action-classifier head and fine-tune the \emph{same}
backbone on the labelled \emph{target} dataset
(either ARID or Dark48) with the KD and CE objective described in
Section \ref{ssec:kd}.
Because the student has never relied on retinex at this stage, inference remains fast and enhancement-free while still inheriting the
teacher’s low-light expertise.

%=====================================================================
\subsection{Stage 2 – Knowledge-Distillation Fine-tuning}
\label{ssec:kd}

Once the student backbone has learned generic low-light features through self-supervision using all clips from ARID and Dark48 with the labels removed, we fine-tune it for action recognition by
\textit{distilling} knowledge from the powerful dual-stream teacher.
The next  stage only dark clips are fed to the student, keeping the model light and enhancement-free, while the teacher
still receives both dark and retinex streams, ensuring its logits
contain the full benefit of Dynamic Feature Fusion and supervised contrastive learning (SupCon). The student copies the teacher’s softened predictions and is nudged by a small cross-entropy term to stay true to the ground-truth label. This lets the student mimic the teacher’s low-light expertise using only dark frames, and keeping the inference fast.

Let $z_t,z_s\in\mathbb{R}^K$ be teacher and student logits. The temperature-scaled (or softened) output distributions of the teacher and student networks are defined as:
\[
p_t^{(\tau)} = \mathrm{softmax}\!\left(\frac{z_t}{\tau}\right), \quad
p_s^{(\tau)} = \mathrm{softmax}\!\left(\frac{z_s}{\tau}\right),
\]
where $\tau > 0$ denotes the temperature parameter that controls the smoothness of the predicted probability distributions.

% We optimize a weighted sum of standard cross-entropy and
% logit-level distillation:
The student network is trained by minimizing a weighted combination of the standard cross-entropy loss and a logit-based knowledge distillation loss:
\begin{align}
\mathcal{L}_{\mathrm{CE}}
&= -\sum_{k=1}^{K} y_k \log \mathrm{softmax}(z_s)_k, \\
\mathcal{L}_{\mathrm{KD}}
&= \tau^{2}\,\mathrm{KL}\!\left(p_t^\tau \,\|\, p_s^\tau\right), \\
\mathcal{L}_{\mathrm{student}}
&= \lambda_{\mathrm{ce}}\mathcal{L}_{\mathrm{CE}}
  + \lambda_{\mathrm{kd}}\mathcal{L}_{\mathrm{KD}} .
\end{align}
% We use $\tau =3$, $\lambda_{\mathrm{ce}}{=}1$,
% and $\lambda_{\mathrm{kd}}=1$.
The temperature parameter is set to $\tau = 4$, and the weighting coefficients are fixed as $\lambda_{\mathrm{ce}} = 1$ and $\lambda_{\mathrm{kd}} = 1$.

%---------------------------------------------------------------------
%---------------------------------------------------------------------

\paragraph{Inference path}
% After fine-tuning, deployment reduces to a \emph{single-stream} forward pass:
After fine-tuning, the inference phase simplifies to a single-stream forward pass:
\[
\text{dark clip only}
\;\longrightarrow\;
g_{\theta}^{\text{KD}}
\;\longrightarrow\;
\text{BERT}
\;\longrightarrow\;
\text{linear head}
\;\longrightarrow\;
\text{action label}.
\]

\begin{equation}
\begin{aligned}
\hat{y}
= \arg\max\;
\sigma\!\bigl(
  W\, h_{\phi}^{\text{KD}}\!\big( g_{\theta}^{\text{KD}}(x^{\text{dark}}) \big) + b
\Big)
\end{aligned}
\end{equation}

\noindent
where  
$\,x^{\text{dark}}\in\mathbb R^{3\times L\times H\times W}$ is the dark clip,  
$g_{\theta}$ is the R(2+1)D-34 backbone (weights fixed after KD), \(h_{\phi}^{\mathrm{KD}}\) is BERT (we use its \([\text{CLS}]\) embedding, weights fixed after KD)
$W\in\mathbb R^{K\times d}$ and $b\in\mathbb R^{K}$ form the linear
action head,  
$\sigma$ denotes softmax, and  
$\hat y$ is the predicted action label.

\section{Experiments}

\subsection{Dataset}

Public benchmarks containing genuine low-light footage are still scarce, so we evaluate our method on the two largest dark-video datasets currently available. ARID~\cite{xu2021ARID} was filmed entirely at night or in severely under-lit indoor scenes and is released in two versions, \ie, V1.0 with 3784 clips and the expanded V1.5 with 5572 clips. Both versions cover the same eleven everyday actions, \ie, drinking, jumping, picking, pouring, pushing, running, sitting, standing, turning, walking, and waving—with each clip lasting between 6 and 10 seconds. The V1.5 release further ensures a minimum of 320 examples per class. To test scalability beyond this limited label set, we also employ the Dark48 dataset \cite{tu2023dtcm}, a collection of 10243 night-time clips taken from surveillance and low-exposure consumer recordings that span 48 action categories over four times the class diversity of the ARID dataset. 

\subsection{Implementation details}
The proposed approach is implemented on an Advanced Computing Platform
(HPC) powered by NVIDIA A100 GPU utilizing the open-source machine learning framework PyTorch~\cite{paszke2019pytorch}. We evaluate the proposed approach on both releases of the ARID benchmark V1.0 and the larger V1.5, as well as on the broader Dark48 dataset, allowing us to measure performance across progressively richer, real-world low-light scenarios. We follow the approach outlined in \cite{chen2021darklight} to report the average Top-1 and Top-5 accuracies across three splits. The input frame dimensions are set to $3 \times 64 \times 112 \times 112$. Both teacher and student use the same backbone, an R(2+1)D-34 \cite{tran2018closer} pretrained on IG-65M \cite{ghadiyaram2019large}, with the final temporal average-pool removed. As for the Teacher (dual-stream, with fusion), we form two synchronized inputs: the original dark clip and a retinex enhanced version \cite{guo2016lime}. Both paths share parameters within the R(2+1)D-34 backbone. Each path produces a feature tensor of $512 \times 8 \times 7 \times 7$. We apply spatial average pooling to obtain a per-timestep sequence $512 \times 8$, transpose to $8 \times 512$, and feed both sequences to a Dynamic Feature Fusion (DFF). We further feed the obtained features into BERT~\cite{devlin2019bert} followed by a basic linear layer to get the model’s final classification result. The teacher is trained with cross-entropy plus supervised contrastive loss. As for the student (single-stream, no fusion), we use only the dark clip as input and the same R(2+1)D-34 backbone. The backbone output $512 \times 8 \times 7 \times 7$ is pooled to $512 \times 8$, transposed to $8 \times 512$, and fed directly to the same BERT-style temporal encoder (no DFF), followed by a basic linear layer to get the model’s final classification result. The student network is trained in two stages: first, through self-supervised pretraining on unlabeled dark clips; and second, through supervised fine-tuning that combines knowledge distillation from the teacher with a cross-entropy objective. During inference, only the student network (single-stream) is employed, resulting in a parameter count identical to that of the backbone (67 M) and eliminating the need for any stream fusion. We trained the student for 50 epochs on ARID V1.0, ARID V1.5, and Dark-48. Unless stated otherwise, we optimize with AdamW (learning rate \(1\times 10^{-5}\)). 
% Student's training proceeds in 2 stages first with a self-supervised pretraining on dark clips (labels removed), followed by supervised fine-tuning with knowledge distillation from the teacher plus cross-entropy. In inference, we only utilize the student (single stream), so parameters match the backbone (67 M), and no fusion is invoked.

\begin{table}[!t]
  \centering
  \caption{Top-1 and Top-5 accuracy results on ARID V1.0 for several competitive methods and our proposed approach}
  \label{tab:accuracy_comparison1_0}
  \setlength{\tabcolsep}{5pt}
  \renewcommand{\arraystretch}{1.1}
  \small

  \begin{threeparttable}
  \begin{tabularx}{\columnwidth}{@{}l
      S[table-format=2.2]  % Top-1
      S[table-format=2.2]  % Top-5
      S[table-format=3.2]  % Params
    @{}}
    \toprule[0.5mm]
    \multicolumn{1}{c}{\textbf{Method (Venue, Year)}} &
    \multicolumn{1}{c}{\textbf{Top-1 (\%)}} &
    \multicolumn{1}{c}{\textbf{Top-5 (\%)}} &
    \multicolumn{1}{c}{\makecell{\textbf{Params}\\\textbf{(M)}}}
\\
    \midrule
    VGG-TS (\textit{ICLR'14}) \cite{simonyan2014very}                  & 32.08 & 90.76 & 144 \\
    TSN (\textit{ECCV'16}) \cite{wang2016temporal}                     & 57.96 & 94.17 & 22  \\
    P3D-199 (\textit{ICCV'17}) \cite{qiu2017learning}                  & 71.93 & 98.66 & 62  \\
    I3D-RGB (\textit{CVPR'17}) \cite{carreira2017quo}                  & 68.29 & 97.69 & 12  \\
    I3D Two-stream (\textit{CVPR'17}) \cite{carreira2017quo}           & 72.78 & 99.39 & 24  \\
    R(2+1)D-34 (\textit{CVPR'18}) \cite{tran2018closer}                & 62.87 & 96.64 & 63  \\
    3D-ResNet-50 (\textit{CVPR'18}) \cite{hara2018can}                 & 71.08 & 99.39 & 46  \\
    3D-ResNet-101 (\textit{CVPR'18}) \cite{hara2018can}                & 71.57 & 99.03 & 86  \\
    DarkLight-ResNeXt-101 (\textit{CVPRW'21}) \cite{chen2021darklight} & 87.27 & 99.47 & 88  \\
    DarkLight-R(2+1)D-34 (\textit{CVPRW'21}) \cite{chen2021darklight}  & 94.04 & 99.87 & 63  \\
    MRAN (\textit{CVPRW'21}) \cite{hira2021delta}                      & 93.73 & \multicolumn{1}{c}{—} & \multicolumn{1}{c}{—} \\
    SCI + R(2+1)D-GCN (\textit{AAAI'23}) \cite{suman2023two}           & 95.86 & 99.87 & \multicolumn{1}{c}{—} \\
    R(2+1)D-GCN+BERT (\textit{TAI'23}) \cite{singh2022action}          & 96.60 & 99.88 & 68  \\
    DTCM (\textit{TIP'23}) \cite{tu2023dtcm}                           & 96.36 & {\bfseries 99.92} & 48 \\
    WiiD\tnote{1} (\textit{ECCV'24}) \cite{li2024watching}             & 58.08 & \multicolumn{1}{c}{—} & \multicolumn{1}{c}{—} \\
    MFDL (\textit{ICASSP'25}) \cite{wang2025advancing}                 & 96.71 & 99.88 & \multicolumn{1}{c}{—} \\
    \midrule[0.25mm]
    \textbf{Proposed: \mymethod (Student)}                        & {\bfseries 96.92} & 99.89 & {\bfseries 67} \\
    \bottomrule[0.5mm]
  \end{tabularx}
  \begin{tablenotes}\footnotesize
    \item[1] Unsupervised-learning setting.
  \end{tablenotes}
  \end{threeparttable}
\end{table}

\begin{table}[!t]
  \centering
  \caption{Top-1 and Top-5 accuracy results on ARID V1.5 for several competitive methods and our proposed approach}
  \label{tab:accuracy_comparison15}
  \setlength{\tabcolsep}{5pt}
  \renewcommand{\arraystretch}{1.1}
  \small
  \begin{threeparttable}
  \begin{tabularx}{\columnwidth}{@{}l
      S[table-format=2.2]   
      S[table-format=2.2]   
      S[table-format=2.0]    
    @{}}
    \toprule[0.5mm]
    \multicolumn{1}{c}{\textbf{Method}} &
    \multicolumn{1}{c}{\textbf{Top-1 (\%)}} &
    \multicolumn{1}{c}{\textbf{Top-5 (\%)}} &
    \multicolumn{1}{c}{\makecell{\textbf{Params}\\\textbf{(M)}}} \\
    \midrule[0.25mm]
    3D-ResNet-18 (\textit{CVPR'17}) \protect\cite{hara2018can}                & 31.16 & 90.49 & 33 \\
    I3D-RGB (\textit{CVPR'17}) \protect\cite{carreira2017quo}                & 48.75 & 90.61 & 12 \\
    I3D Two-stream (\textit{CVPR'17}) \protect\cite{carreira2017quo}         & 51.24 & 90.95 & 24 \\
    DarkLight R(2+1)D-34 (\textit{CVPRW'21}) \protect\cite{chen2021darklight} & 84.13 & 97.34 & 63 \\
    R(2+1)D-GCN + BERT (\textit{TAI'23}) \protect\cite{singh2022action}  & 86.93 & {\bfseries 99.35} & 68 \\
    \midrule[0.25mm]
    \textbf{Proposed: \mymethod (Student)}                               & {\bfseries 88.27} & 99.12 & {\bfseries 67} \\
    \bottomrule[0.5mm]
  \end{tabularx}
  \end{threeparttable}
\end{table}

\begin{table}[!t]
  \centering
  \caption{Top-1 and Top-5 accuracy results on Dark48 for several competitive methods and our proposed approach}
  \label{tab:accuracy_comparisonDark48}
  \setlength{\tabcolsep}{5pt}
  \renewcommand{\arraystretch}{1.1}
  \small
  \begin{threeparttable}
  \begin{tabularx}{\columnwidth}{@{}l
      S[table-format=2.2]      % Top-1
      S[table-format=2.2]      % Top-5
      S[table-format=3.2]      % Params
    @{}}
    \toprule[0.5mm]
    \multicolumn{1}{c}{\textbf{Method}} &
    \multicolumn{1}{c}{\textbf{Top-1 (\%)}} &
    \multicolumn{1}{c}{\textbf{Top-5 (\%)}} &
    \multicolumn{1}{c}{\makecell{\textbf{Params}\\\textbf{(M)}}} \\
    \midrule[0.25mm]
    I3D-RGB (\textit{CVPR'18}) \protect\cite{carreira2017quo}                  & 32.25 & 65.35 & 12.29 \\
    3D-ResNet-50 (\textit{CVPR'18}) \protect\cite{hara2018can}                & 34.26 & 66.82 & 46.22 \\
    3D-ResNet-101 (\textit{CVPR'18}) \protect\cite{hara2018can}                & 36.11 & 68.74 & 85.27 \\
    DarkLight-R(2+1)D-34 (\textit{CVPRW'21}) \protect\cite{chen2021darklight}  & 39.08 & 71.24 & 66.73 \\
    DarkLight-ResNeXt-101 (\textit{CVPRW'21}) \protect\cite{chen2021darklight} & 42.27 & 70.47 & 88.00 \\
    DTCM (\textit{TIP'23}) \protect\cite{tu2023dtcm}                           & 46.68 & 75.92 & 47.57 \\
    MFDL (\textit{ICASSP'25}) \protect\cite{wang2025advancing}                 & 48.14 & 76.67 & \multicolumn{1}{c}{—} \\
    \midrule[0.25mm]
    \textbf{Proposed: \mymethod (Student)}                                & {\bfseries 48.96} & {\bfseries 76.71} & {\bfseries 67} \\
    \bottomrule[0.5mm]
  \end{tabularx}
  \end{threeparttable}
\end{table}

\subsection{Results and Discussion}
We compare the performance of our method with several state-of-the-art approaches for action recognition in dark videos, as presented in Tables~\ref{tab:accuracy_comparison1_0}, \ref{tab:accuracy_comparison15}, and \ref{tab:accuracy_comparisonDark48}. These results include evaluations on both versions of the ARID dataset, which includes the V1.0 and V1.5, as well as the Dark48 dataset. Most of the data are sourced from \cite{singh2022action}. Our method sets a new state-of-the-art on both ARID versions (V1.0 and V1.5) and on Dark48. All eleven ARID classes and all forty-eight Dark48 classes are evaluated, with performance reported in Top-1 and Top-5 accuracy. All reported numbers are obtained using our student network, which employs an R(2+1)D-34 backbone trained via self-supervised learning on dark frames, followed by knowledge distillation from a more capable teacher model.

%The results of our method, along with those of current competitive methods for action recognition in dark videos, are presented in Table \ref{tab:accuracy_comparison1_0}, Table \ref{tab:accuracy_comparison15}, and Table \ref{tab:accuracy_comparisonDark48}. 

Although the student and the teacher share the same backbone architecture, they differ fundamentally in design and learning capacity. The teacher operates in a multi-stream setting, processing both dark and retinex enhanced frames, and incorporates a Dynamic Feature Fusion (DFF) module to adaptively emphasize the most informative temporal segments. Additionally, the teacher is trained with a supervised contrastive loss that sharpens class boundaries by explicitly optimizing feature separability. In contrast, the student receives only a single dark-frame stream and omits DFF and contrastive learning, making its dark training less informative. 

Knowledge distillation bridges the supervision gap between the multi-stream teacher and the single-stream student. The student is trained on the teacher’s temperature-scaled targets, which allows the student to absorb the teacher’s spatio-temporal cues and context-aware weighting. Despite seeing only a single dark-frame stream and using a simpler architecture, it is able to close the gap of the teacher. However, with self-superivsed learning the student is able to marginally surprass the teacher. Hence, KD is an essential mechanism that allows a single-stream model to rival or exceed multi-stream baselines, while remaining substantially lighter and faster at inference.
% Hence, KD is an essential mechanism that allows a single-stream model to rival or exceed multi-stream baselines, while remaining substantially lighter and faster at inference.

\mymethod~achieves 96.92\% Top-1 accuracy on ARID V1.0, surpassing the closest competitor, MFDL (96.71\%), by 0.21\% as shown in Table \ref{tab:accuracy_comparison1_0}. Though small, this improvement is significant given the high accuracy baseline, where marginal gains are hard to achieve. Our approach also surpass DTCM by 0.58\% (96.36\%) and R(2+1)D-GCN+BERT by 0.32\% (96.60\%). While DTCM does edge us on Top-5 accuracy (99.92\% vs. 99.89\%), the 0.03\% gap is negligible relative to our superior Top-1 performance. \mymethod~not only tops DarkLight-R(2 + 1)D-34 and SCI + R(2 + 1)D-GCN by 3.1\% and 1.1\%, respectively, but also outperforms the recent unsupervised WiiD framework by a massive 38.8\% (96.92\% vs 58.08\%). Although DarkLight, SCI + R(2+1)D-GCN, and MFDL all employ dual-stream architectures, our dual-stream teacher model achieves superior accuracy. This improvement can be attributed to the synergistic effect of the Dynamic Feature Fusion (DFF) module, which adaptively balances dark and enhanced inputs across temporal segments, and the integration of Supervised Contrastive Learning (SupCon), which strengthens intra-class cohesion and enhances inter-class separation under varying illumination conditions. 

More importantly, after distillation, our student operating with only dark-frame inputs at inference retains these advantages and even matches or surpasses multi-stream baselines, while being significantly more efficient. We also list classic architectures such as VGG-TS, TSN, I3D, 3D-ResNet, and P3D to give a lower-bound reference. These networks were trained on bright-light datasets and lack any built-in low-light handling, so their lower scores show the drop in performance when a daylight model is used unchanged on dark videos. In contrast, \mymethod~is specifically designed for dark videos; therefore, it surpasses these baselines by a huge margin.
% Despite DarkLight, SCI + R(2+1)D-GCN, and MFDL all relying on dual-stream architectures, our dual-stream teacher itself achieves superior accuracy, owing to the combined strength of our (DFF) module, which adaptively balances dark and enhanced inputs at each temporal segment, and the use of Supervised Contrastive Learning (SupCon,) which tightens intra-class clusters and sharpens inter-class margins under lighting variation

Table \ref{tab:accuracy_comparison15} shows that legacy backbones trained for well-lit videos (3D-ResNet-18, I3D-RGB, and I3D two-stream) collapse in near-dark scenes, topping out at 51.24\% Top-1 despite ranging from 12 M to 33 M parameters. Introducing low-light–specific tricks yields large gains, which include DarkLight R(2+1)D-34 that climbs to 84.13\% and R(2+1)D-GCN+BERT pushes to 86.93\%. Our distilled R(2+1)D-34 student achieves 88.27\% Top-1 and 99.12\% Top-5 accuracy, surpassing the previous best by 1.34\% in Top-1 and outperforming DarkLight by a substantial margin of 4.14\%. Notably, this is accomplished using only a single dark-frame stream at inference, whereas DarkLight relies on a dual-stream setup incorporating both dark and gamma-enhanced inputs.

Dark48 is widely regarded as one of the most demanding low-light benchmarks. Table \ref{tab:accuracy_comparisonDark48} shows that conventional backbones, which include I3D-RGB, 3D-ResNet-50, and 3D-ResNet-101, perform poorly under low light, showing that a model trained for daylight loses a lot of accuracy when used unchanged. Even low-light–aware baselines fall short, DarkLight-R(2+1)D-34 reaches (39.08\%) in Top-1, DarkLight-ResNeXt-101 has reached (42.27\%), and DTCM has achieved (46.68\%). MFDL, the previous best-performing approach, attains 48.14\% Top-1 accuracy. In comparison, \mymethod~ advances the state-of-the-art, achieving 48.96\% Top-1 and 76.71\% Top-5 accuracy surpassing MFDL by 0.91\% in Top-1 and 0.04\% in Top-5 performance, despite employing a simpler single-stream input design. Our method outperforms DarkLight-R(2+1)D-34 by a huge margin, primarily because the teacher’s Dynamic Feature Fusion (DFF) module, whose benefits are transferred to the student through knowledge distillation, dynamically re-weights the two streams at each time step, emphasizing the most informative temporal segments. In contrast, DarkLight employs a fixed concatenation strategy, which can underutilize complementary cues.
% MFDL, the previous best performing approach, climbs to 48.14\% Top-1 accuracy. \mymethod~raises the bar to 48.96\% Top-1 and 76.71\% Top-5, edging past MFDL by 0.91\% Top-1 and by 0.04\% in Top-5 despite using a simpler single-stream input.

\subsection{Ablation Study}

\begin{table}[!t]
  \centering
  \caption{Ablation on the knowledge distillation (KD) pipeline on ARID V1.0, ARID V1.5, and Dark48.}
  \label{table:ablation_kd}
  \renewcommand{\arraystretch}{1.1}
  \scriptsize
  \resizebox{\columnwidth}{!}{%
  \begin{tabular}{@{}l c rr rr rr @{}}
    \toprule
    \textbf{Configuration} & \textbf{Params (M)}
    & \multicolumn{2}{c}{\textbf{ARID V1.0}}
    & \multicolumn{2}{c}{\textbf{ARID V1.5}}
    & \multicolumn{2}{c}{\textbf{Dark48}} \\
    & & \textbf{Top-1} & \textbf{Top-5}
      & \textbf{Top-1} & \textbf{Top-5}
      & \textbf{Top-1} & \textbf{Top-5} \\
    \midrule
    Teacher: DFF + SupCon                  & 69 & 96.79 & 99.91 & 87.53 & 98.56 & 48.42 & 75.99 \\
    Student: SSL only (no KD)              & 67 & 91.11 & 98.21 & 82.15 & 96.56 & 43.21 & 71.98 \\
    \textbf{Student: SSL + KD (\mymethod)} & \textbf{67} & \textbf{96.92} & \textbf{99.89} & \textbf{88.27} & \textbf{99.12} & \textbf{48.96} & \textbf{76.71} \\
    \bottomrule
  \end{tabular}}
\end{table}

This section presents an ablation study of the blocks used in the proposed \mymethod~architecture. Table \ref{table:ablation_kd} makes clear that knowledge distillation (KD), when combined with our self-supervised learning (SSL), is the decisive factor in closing and even marginally beating the performance of the teacher. Compared to the SSL-only student, adding KD yields substantial Top-1 gains of 5.81\%, 6.12\%, and 5.75\% on ARID V1.0, ARID V1.5, and Dark48, respectively, alongside notable Top-5 improvements. More importantly, the distilled single-stream surpasses the multi-stream teacher across all three benchmarks, 0.13\% Top-1 on ARID V1.0, 0.74\% on ARID V1.5,  and 0.54\% on Dark48 while avoiding the teacher’s multi-stream inference overhead. These gains arise because KD transfers knowledge from the teacher’s dual-stream DFF and SupCon pipeline into the student’s single-stream representation. Through soft targets, the student inherits the teacher’s ability to weight temporal segments adaptively and preserve fine-grained inter-class structure without direct access to the retinex enhanced frames. SSL complements this by embedding low-light–robust invariances, enabling the student to generalize more effectively to low-light scenario. The end result is a single-stream recognizer with multi-stream-level accuracy and lower inference cost. It is worth noting that the teacher model alone surpasses the previous best results on ARID V1.0 and ARID V1.5, and achieves performance comparable to the previous best model on Dark48. This highlights the effectiveness of the proposed DFF and supervised contrastive learning components in enhancing recognition accuracy under low-light conditions.

\begin{table}[!t]
\centering
\caption{Impact of fusion strategies on the teacher and the effect of KD on the student across ARID V1.0, ARID V1.5, and Dark48.}
\label{table:dff_joint_stacked}
\renewcommand{\arraystretch}{1.12}
\scriptsize
\resizebox{\columnwidth}{!}{%
\begin{tabular}{@{}l c rr rr rr @{}}
\toprule
\textbf{Fusion Strategy} & \textbf{Params (M)}
& \multicolumn{2}{c}{\textbf{ARID V1.0}}
& \multicolumn{2}{c}{\textbf{ARID V1.5}}
& \multicolumn{2}{c}{\textbf{Dark48}} \\
& & \textbf{Top-1} & \textbf{Top-5}
  & \textbf{Top-1} & \textbf{Top-5}
  & \textbf{Top-1} & \textbf{Top-5} \\
\midrule
\multicolumn{8}{@{}c@{}}{\textbf{Teacher (fusion variant) + SupCon}} \\
\midrule
No fusion (single stream: dark only)       & 67.5 & 90.57 & 98.72 & 82.91 & 96.52 & 42.95 & 70.83 \\
No fusion (single stream: retinex only)    & 67.5 & 92.51 & 99.81 & 85.73 & 98.11 & 46.15 & 73.82 \\
Static fusion (concatenation)              & 68.0 & 94.82 & 99.85 & 86.11 & 98.34 & 48.02 & 75.91 \\
\textbf{DFF (dynamic re-weighting, ours)}  & \textbf{69.0} & \textbf{96.79} & \textbf{99.91} & \textbf{87.53} & \textbf{98.56} & \textbf{48.42} & \textbf{75.99} \\
\midrule
\multicolumn{8}{@{}c@{}}{\textbf{Student (SSL + KD) distilled from corresponding teacher}} \\
\midrule
Teacher\textemdash{}dark only              & 67.0 & 91.61 & 99.05 & 83.47 & 97.15 & 43.07 & 71.17 \\
Teacher\textemdash{}retinex only           & 67.0 & 93.02 & 99.79 & 85.89 & 98.55 & 46.77 & 73.95 \\
Teacher\textemdash{}static fusion          & 67.0 & 95.88 & 99.82 & 86.35 & 98.45 & 48.32 & 76.07 \\
\textbf{Teacher\textemdash{}DFF (ours)}    & \textbf{67.0} & \textbf{96.92} & \textbf{99.89} & \textbf{88.27} & \textbf{99.12} & \textbf{48.96} & \textbf{76.71} \\
\bottomrule
\end{tabular}}
\end{table}

The results in Table \ref{table:dff_joint_stacked} highlight two critical findings: the superiority of dynamic feature fusion (DFF) over simpler fusion strategies in the teacher model, and the ability of knowledge distillation (KD) with self-supervised learning (SSL) to transfer that advantage to a single-stream student. Among the teacher configurations, the single-stream baselines using only dark or only retinex-enhanced inputs perform noticeably worse, reinforcing that effective low-light action recognition relies on exploiting complementary information from both streams. Static fusion via concatenation improves over single-stream setups but still treats the streams equally, failing to adapt to changes in the relative importance of each stream over time. By contrast, DFF adaptively re-weights the two streams at each time step and achieves the highest Top-1 accuracy across ARID V1.0, V1.5, and Dark48, providing clear evidence that dynamic, context-dependent fusion better exploits complementary information. Crucially, distilling the DFF-enhanced multi-stream teacher into our SSL-pretrained single-stream student improves accuracy on all 3 of the datasets, largest on ARID V1.5 and Dark48 (+0.74\% and +0.54\% Top-1), while reducing inference to a single dark stream with a fraction of the teacher’s input complexity. This works because KD conveys more than labels, through the teacher’s soft targets, it transfers the effect of DFF’s content-aware, per-timestep fusion, allowing the student to approximate multi-stream reasoning from dark frames alone. Combined with SSL, this yields a single-stream recognizer that delivers multi-stream-level accuracy at a fraction of the inference cost.

\begin{table}[!t]
\centering
\caption{Ablation of self-supervised pretraining (SSL) variants for the single-stream student (KD fixed). Two dark views are used only for SSL pretraining; inference is single-stream. Backbone: R(2+1)D-34.}
\label{tab:ssl_spatial_temporal_ablation_filled}
\renewcommand{\arraystretch}{1.12}
\scriptsize
\resizebox{\columnwidth}{!}{%
\begin{tabular}{@{}l c rr rr rr @{}}
\toprule
\textbf{Student variant} & \textbf{Params (M)}
& \multicolumn{2}{c}{\textbf{ARID V1.0}}
& \multicolumn{2}{c}{\textbf{ARID V1.5}}
& \multicolumn{2}{c}{\textbf{Dark48}} \\
& & \textbf{Top-1} & \textbf{Top-5}
  & \textbf{Top-1} & \textbf{Top-5}
  & \textbf{Top-1} & \textbf{Top-5} \\
\midrule
Student (KD only; no SSL)                    & 67 & 95.75 & 99.85 & 87.36 & 98.95 & 47.99 & 75.80 \\
SSL \textemdash{} spatial only               & 67 & 96.10 & 99.76 & 87.41 & 98.98 & 48.05 & 76.10 \\
SSL \textemdash{} temporal only              & 67 & 96.35 & 99.85 & 87.80 & 99.00 & 48.40 & 76.30 \\
\textbf{SSL \textemdash{} spatial + temporal (ours)} 
                                             & \textbf{67} & \textbf{96.92} & \textbf{99.89} & \textbf{88.27} & \textbf{99.12} & \textbf{48.96} & \textbf{76.71} \\
\bottomrule
\end{tabular}}
\end{table}

The ablation in Table \ref{tab:ssl_spatial_temporal_ablation_filled} shows that adding SSL on top of a fixed KD recipe yields consistent gains. Relative to KD-only, spatio-temporal SSL improves Top-1 by 1.17\% on ARID V1.0, 1.17\%  on ARID V1.5, and 1.26\% on Dark48, with modest Top-5 lifts (ARID is already near saturation). The results indicate that spatio-temporal augmentations are much better than using temporal and spatial augmentations individually across all three datasets. Moreover, temporal-only augmentations outperform spatial-only by 0.70\% vs 0.35\% Top-1 over KD on V1.5, which is expected in low-light video where motion cues carry more signal than appearance. Crucially, all rows use the same 67 M backbone and identical KD approach, so the gains come purely from SSL; inference remains single-stream and unchanged in cost. It is worth noting that the teacher’s results are reported in Table \ref{table:ablation_kd} and Table \ref{table:dff_joint_stacked} shows that the dual-stream teacher reaches 96.79\%/ 87.53\%/ 48.42\% Top-1 on ARID V1.0 / ARID V1.5 / Dark48, respectively. As expected, the KD-only student in Table \ref{tab:ssl_spatial_temporal_ablation_filled} trails this teacher, 95.75\% /87.10\% /47.70\%. Although distillation transfers the DFF-induced, SupCon-sharpened decision patterns, the student has access to only dark frames and no explicit DFF/SupCon mechanism, so it can only approximate the teacher’s multi-stream reasoning. Crucially, adding SSL (still single-stream at inference) closes the gap and slightly surpasses the teacher on all three benchmarks 96.92 / 88.27 / 48.96 without changing parameters (67M) or inference cost. This shows that SSL provides complementary information beyond KD, leading to stronger generalization.

\begin{figure}[t]
\centering
\includegraphics[width=0.9\textwidth]{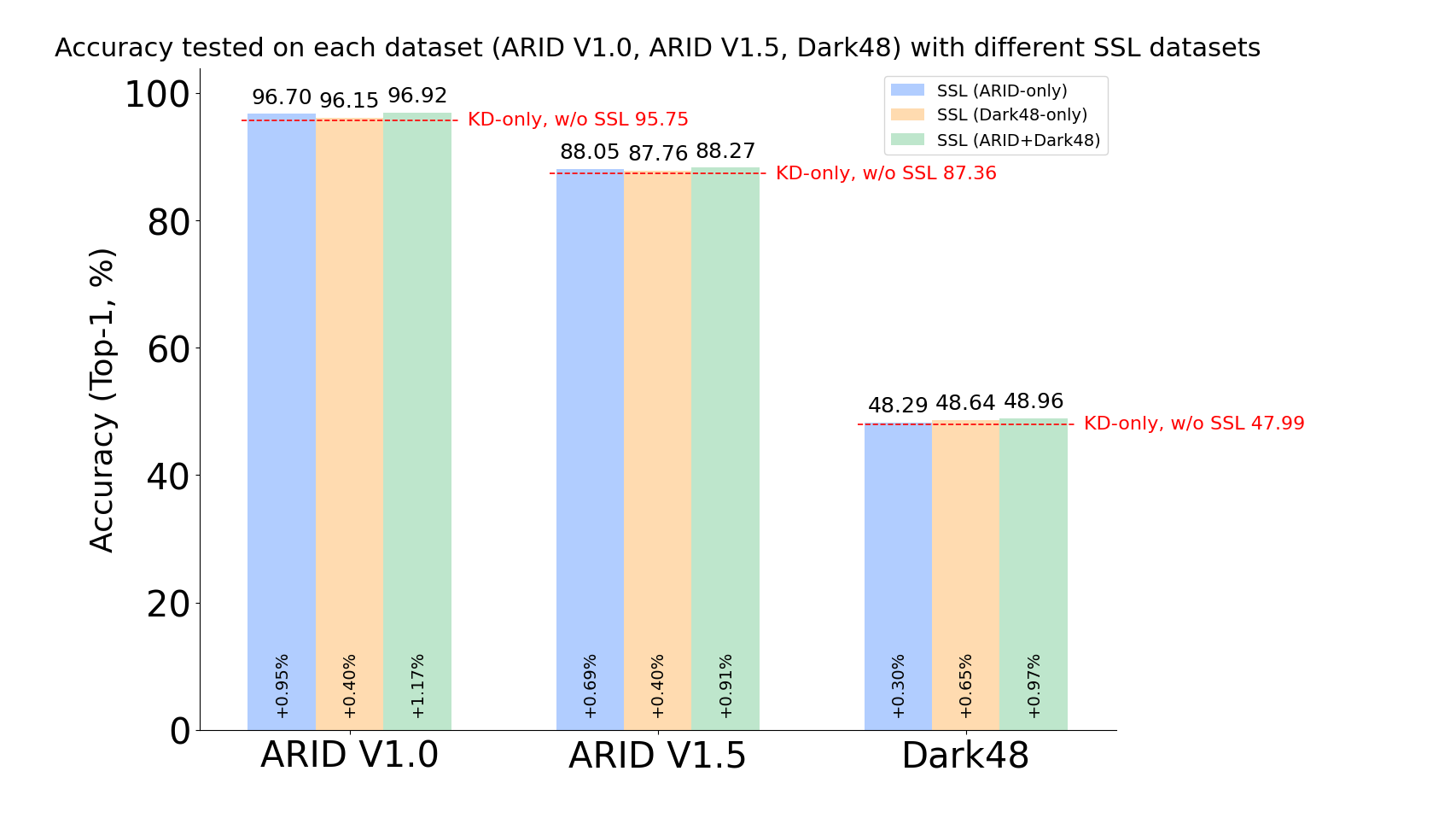}
\caption{Effect of unlabeled SSL pretraining source on downstream \textbf{Top-1} accuracy for \textbf{ARID V1.0}, \textbf{ARID V1.5}, and \textbf{Dark48}. Each group compares SSL on \emph{ARID-only}, \emph{Dark48-only}, and \emph{ARID+Dark48 (combined)}. Red dashed lines denote the \emph{KD-only} (no SSL) baseline for that dataset. Numbers above bars show absolute accuracy, with the improvement over KD-only shown at the bottom (near the x-axis) of the chart. Combined pretraining is best across all datasets, and in-domain SSL consistently outperforms cross-domain SSL.}

\label{fig.ssl}
\end{figure}

Figure \ref{fig.ssl} shows that SSL pretraining consistently improves over the KD-only student across all three datasets for each pretraining source (ARID-only, Dark48-only, and ARID+Dark48). Two clear patterns emerge: the first is that domain match matters; in-domain SSL (ARID→ARID, Dark48→Dark48) beats cross-domain in every case, indicating that dataset-specific low-light priors transfer best. Besides that, diversity helps most; the combined unlabeled pool (ARID+Dark48) is strongest on all benchmarks, suggesting complementary coverage regularizes the representation and yields invariances that KD can exploit. Absolute Top-1 gains are largest on ARID V1.0 (+1.17\%), while the largest relative gain appears on Dark48 (+0.97\% on a 47.99 baseline $\approx$ +2.0\%), reflecting greater headroom on the harder, 48-class dataset. Notably, cross-domain SSL never hurts; gains remain positive, showing that spatiotemporal cues learned from dark video are broadly transferable.

\begin{table}[!t]
\centering
\caption{Teacher ablation with DFF fixed: with vs.\ without SupCon.}
\label{tab:teacher_supcon_ablation}
\renewcommand{\arraystretch}{1.12}
\scriptsize
\resizebox{\columnwidth}{!}{%
\begin{tabular}{@{}l c rr rr rr @{}}
\toprule
\textbf{Configuration (Teacher)} & \textbf{Params (M)}
& \multicolumn{2}{c}{\textbf{ARID V1.0}}
& \multicolumn{2}{c}{\textbf{ARID V1.5}}
& \multicolumn{2}{c}{\textbf{Dark48}} \\
& & \textbf{Top-1} & \textbf{Top-5}
  & \textbf{Top-1} & \textbf{Top-5}
  & \textbf{Top-1} & \textbf{Top-5} \\
\midrule
DFF teacher, \textit{without} SupCon & 69.0 & 96.30 & 99.86 & 86.95 & 98.34 & 47.72 & 75.60 \\
\textbf{DFF teacher, \textit{with} SupCon} & \textbf{69.0} & \textbf{96.79} & \textbf{99.91} & \textbf{87.53} & \textbf{98.56} & \textbf{48.42} & \textbf{75.99} \\
\bottomrule
\end{tabular}}
\begin{minipage}{\columnwidth}\vspace{4pt}\footnotesize
All settings identical across rows; DFF is enabled in both.
\end{minipage}
\end{table}

Table \ref{tab:teacher_supcon_ablation} shows that the supervised contrastive term consistently boosts the DFF teacher across all datasets at fixed capacity (69 M) and training schedule. Relative to a DFF-only teacher, adding SupCon improves Top-1 by +0.49\% on ARID V1.0, +0.58\% on ARID V1.5, and +0.7\% on Dark48 with parallel Top-5 gains. The effect is largest on Dark48 (+0.7\%) due to the fact that Dark48 has 48 classes, where denser, fine-grained class boundaries make margin sharpening more impactful. By contrast, ARID has only 11 classes, so inter-class overlap is lower and the incremental benefit is correspondingly smaller. SupCon still tightens clusters and separates neighbors, producing cleaner per-timestep features that DFF fuses more effectively.

%\begin{align}
%\mathbf{p}_s^\tau &= \operatorname{softmax}(\mathbf{z}_s/\tau), \qquad
%\mathbf{p}_t^\tau = \operatorname{softmax}(\mathbf{z}_t/\tau),\\
%\mathcal{L}_{\text{CE}} &= -\sum_{k=1}^{K} y_k \log \operatorname{softmax}(z_{s})_k,\\
%\mathcal{L}_{\text{KD}} &= \tau^{2}\,\mathrm{KL}\!\left(\mathbf{p}_t^\tau \,\|\, \mathbf{p}_s^\tau\right),\\
%\mathcal{L}_{\text{student}} &= \lambda_{\text{ce}}\,\mathcal{L}_{\text{CE}}
%+ \lambda_{\text{kd}}\,\mathcal{L}_{\text{KD}}.
%\end{align}

\section{Conclusion}

We introduced \mymethod, a teacher–student framework for dark video action recognition. The dual-stream teacher uses Dynamic Feature Fusion (DFF) to adaptively re-weight dark and retinex features at each temporal segment, while supervised contrastive learning (SupCon) sharpens inter-class margins. The student is single-stream and is SSL-pretrained to learn low-light invariances and improve generalization from unlabeled video, then fine-tuned with KD. Through distillation, the teacher’s multi-stream cues and stronger decision boundaries are transferred to a lightweight model that requires only one stream at inference. The resulting framework sets a new state-of-the-art with single-stream inference: 96.92\% / 99.89\% (Top-1/Top-5) on ARID V1.0, 88.27\%  /99.12\%  on ARID V1.5, and 48.96\% / 76.71\%  on Dark48, combining strong accuracy with practical efficiency. Ablation studies confirm additive contributions from each component. Future work includes scaling SSL to larger unlabeled corpora and extending \mymethod~to other adverse conditions.

\bibliographystyle{elsarticle-num} % or elsarticle-num-names
\bibliography{cas-refs}              
\end{document}